\documentclass[conference]{ieeeconf}

\IEEEoverridecommandlockouts

\usepackage{graphicx}        

\usepackage[cmex10]{amsmath}
\usepackage{amssymb}

\usepackage{cite}

\usepackage{comment}
\usepackage{verbatim}
\usepackage{subfigure}
\usepackage{wrapfig}
\usepackage{url}
\usepackage{color}
\newcommand{\red}[1]{\textcolor{red}{#1}}

\DeclareMathAlphabet{\mathcal}{OMS}{cmsy}{m}{n}

\makeindex

\begin{document}
\bstctlcite{IEEEexample:BSTcontrol}
\title{\LARGE \bf 3-D Reciprocal Collision Avoidance on Physical Quadrotor Helicopters with On-Board Sensing for Relative Positioning}
\author{Parker Conroy  \and Daman Bareiss \and Matt Beall \and Jur van den Berg%
\thanks{Parker Conroy,Daman Bareiss, and Matt Beall are with the Department of Mechanical Engineering at the University of Utah. E-mail: parcon@cs.utah.com, daman.bareiss@utah.edu, matt.beall@utah.edu.}
\thanks{Jur van den Berg is with the School of Computing at the University of Utah. E-mail: berg@cs.utah.edu.} 
}

\maketitle

\begin{abstract}
In this paper, we present an implementation of 3-D reciprocal collision avoidance on real quadrotor helicopters where each quadrotor senses the relative position and velocity of other quadrotors using an on-board camera. We show that using our approach, quadrotors are able to successfully avoid pairwise collisions GPS and motion-capture denied environments, without communication between the quadrotors, and even when human operators deliberately attempt to induce collision. To our knowledge, this is the first time that reciprocal collision avoidance has been successfully implemented on real robots where each agent independently observes the others using on-board sensors. We theoretically analyze the response of the collision-avoidance algorithm to the violated assumptions by the use of real robots. We quantitatively analyze our experimental results. A particularly striking observation is that at times the quadrotors exhibit ``reciprocal dance'' behavior, which is also observed when humans move past each other in constrained environments. This seems to be the result of sensing uncertainty, which causes both robots involved to have a different belief about the relative positions and velocities and, as a result, choose the same side on which to pass.   
 

\end{abstract}

\section{Introduction}
Reciprocal collision avoidance (RCA) has been actively studied in robotics over the past few years. The problem can generally be defined in the context of multiple autonomous mobile robots navigating a common environment, where each robot employs a decentralized  continuous sensing-control cycle. In each cycle, each robot must independently compute an action based on its local observations of the other robots, without mutual communication or coordination, such that it stays free of collisions with the other robots and progresses along a user controlled trajectory. The key aspect of most reciprocal collision avoidance approaches is that they specifically account for the reactive nature of the other robots, assuming each robot takes half the responsibility of avoiding pairwise collisions. Failing to do so would inherently cause undesirable oscillations in the motion of the robots \cite{rvo}. Basic reciprocal collision avoidance approaches apply to robots with simple holonomic dynamics \cite{rvo, nbody}, but more recently, RCA has been extended to robots with differential-drive dynamics \cite{snape2010,alonso2010optimal}, car-like dynamics \cite{alonso2012reciprocal}, double-integrator dynamics \cite{avo, lalish2012distributed}, arbitrary-degree integrator dynamics \cite{Ruffli2013}, and general linear dynamics \cite{Bareiss_2013}.

An appealing feature of optimal reciprocal collision avoidance (ORCA) \cite{nbody} is that it is based on the \emph{relative velocity} paradigm \cite{fiorini1993motion}. That is, in addition to the absolute velocity of oneself, each of the robots only needs information about the \emph{relative} position and the \emph{relative} velocity of other robots; each robot does not need to estimate the full state of the other robot. Gathering this information is still not trivial, but is much simpler than gathering the opposite robot's full global state. In principle, this makes the approach well-suited for use in environments where absolute position information from e.g. GPS or a motion capture system is unavailable, as the robots can acquire relative velocity information from on-board sensors. Most previous implementations of RCA, however, have only been applied in simulation.

Applying RCA on real robots poses a number of unique challenges. In particular, the ORCA framework is built on the assumption of perfect symmetry (i.e. a pair of robots observe exactly the same relative position and velocity with respect to each other) and perfect reciprocity (i.e. both robots take half the responsibility of avoiding collisions) to guarantee collision avoidance. In settings where each robot uses its own on-board sensing, the symmetry assumption is inherently violated due to sensor noise. In addition, the ORCA framework assumes that robots can adopt any velocity instantaneously. While extensions exist for robots with more complicated dynamics, real robots will always deviate from the expected behavior due to external disturbances or modeling error.

This paper explores the effect of the violation of these assumptions on the behavior of a 3-D implementation of ORCA \cite{nbody} in GPS and motion capture denied environments on fully independent quadrotor helicopters that sense other robots using their own on-board sensors. To our knowledge, this paper is the first to apply reciprocal collision-avoidance on real robots using on-board decentralized sensing.

\begin{figure}
\centering
\includegraphics[width = \columnwidth]{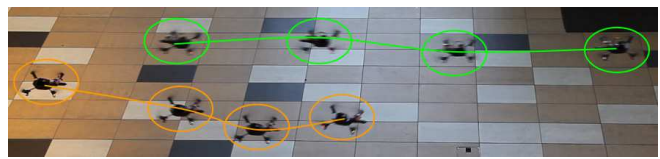}
\caption{Two quadrotors are controlled along a straight-line trajectory by a user and our reciprocal collision avoidance algorithm corrected the control velocities to produce a collision-free trajectory.}
\label{Fig:intro}
\vspace{-10pt}
\end{figure}

We both qualitatively and quantitatively analyze the effect the violation of symmetry, reciprocity, and dynamic assumptions has on the collision avoidance behavior of the robots, and we show that ORCA is mostly robust against these violations. In particular, our experimental results suggest that sensor noise can lead to \emph{reciprocal dances}, even though not leading to collisions. This phenomenon is also observed in human motion. Our analysis suggests this is the result of asymmetry in the sensing of relative position and velocity between robots. Each agent believes the optimal avoidance path is to the same side. The result is that both robots choose a trajectory that is anew on a collision-course, which may lead to a repetition of the phenomenon until the sensing asymmetry breaks. This phenomenon is most likely when the robots approach each other ``head-on'', in which case there is no obvious side on which to pass. We will theoretically discuss the nature and origin of reciprocal dances in RCA.

Also, our experimental results suggest that ORCA is robust against the violation of the assumed dynamics. The robots will exponentially fast converge to a collision free trajectory even when the robots are not able to instantaneously assume such a trajectory. The same mechanism is at play when the reciprocity assumption, i.e. the assumption that the other robot takes care of its share of the responsibility of avoiding collisions, is violated by an other robot. The robot will then converge to a collision avoiding trajectory with an exponential rate. With our approach being reactive, it can respond to other robots, previously untracked, suddenly appearing.


We experimented with our framework on a pair of Parrot AR.Drone quadrotors in multiple environments where no external motion capture or GPS sensing were used. All the necessary sensing was preformed by on-board CMOS cameras. On board vision algorithms were used to detect other quadrotors in the image frame of a forward-facing camera. As the crafts entered each other's detection envelope, these noisy measurements were processed through a Kalman filter which estimated the relative position and relative velocity with respect to other quadrotors. In all of our experiments, the quadrotors were flown by human operators, and no collisions occured even if the operators attempted to steer the quadrotors along malicious, colliding trajectories.


The paper is organized as follows: An overview of the components concerning the 3-D ORCA algorithm are given in Section \ref{orca}. On-board sensing and the details of our implementation are discussed in Section \ref{sensing}. The experimental framework is reviewed in Section \ref{setup}.  Conclusions are made in Section \ref{results}.

\section{Previous Work} 
The concept of reciprocal collision avoidance was first introduced in \cite{rvo}, which presented Reciprocal Velocity Obstacles as an extension to Velocity Obstacles (VO) \cite{fiorini1998motion}, where agents actively attempt to avoid collisions with each other. While this approach overcame the oscillations observed with VO and guaranteed collision avoidance for a pair of robots, it still exhibited oscillations in settings with more than two robots. Hybrid RVO \cite{snape2009independent, snape2011hybrid} was introduced to mitigate this undesired behavior, but does not offer any formal guarantees on smoothness and collision avoidance. Optimal Reciprocal Collision Avoidance (ORCA) \cite{nbody} addressed these issues, and formal proofs of smoothness and collision avoidance for an arbitrary number of robots were given in \cite{snape2010}.

The aforementioned approaches focus on robots with idealized, holonomic dynamics, i.e. robots that can instantaneously adopt any velocity in the 2-D plane. While this assumption can be applied to differential-drive robots by enlarging the effective robot radius \cite{snape2010}, it does not generalize to robots with more involved dynamics. Acceleration-velocity obstacles (AVO) \cite{avo} partly overcome this, and guarantee collision avoidance for robots with double-integrator dynamics (i.e. omni-directional acceleration control). This formulation can be applied to robots with car-like dynamics using a change of variables in the dynamics. Alonso-Mora et al. provide more direct formulations for reciprocal collision avoidance of robots with car-like dynamics \cite{alonso2010optimal, alonso2012reciprocal}. Recently, more general approaches for reciprocal collision avoidance have been introduced: The approach in \cite{Ruffli2013} generalizes AVO and provides a formulation for robots with arbitrary-degree integrator dynamics (i.e. omnidirectional control of acceleration, jerk, snap, etc.). Reciprocal LQR-Obstacles \cite{Bareiss_2013} allow for collision avoidance of robots with arbitrary linear dynamics. Another centralized method developed to be able to handle suddenly appearing or pop-up obstacles was introduced by \cite{sujit2009multiple} but requires on-the-fly computation of the robot's full trajectory.

While the above approaches have successfully been tested in simulation, and some have been applied to collision avoidance for agents in virtual environments and games \cite{space_marine}, few reciprocal collision approaches have been applied on real robots. HRVO, ORCA, and AVO \cite{snape2011hybrid,snape2010,snape2010iser} have been applied to iRobot Create robots with differential-drive dynamics, and ORCA has been applied to quadrotors \cite{mora12object}, but with centralized sensing where a single sensor observes the environment and broadcasts the observed robot positions and velocities to all robots. To the knowledge of the authors, neither 2-D or 3-D RCA has been experimentally verified on independent agents with distributed, on-board sensing, where each robot perceives its own surroundings. 

Other, mostly centralized approaches to collision avoidance, i.e. approaches where motions of robots are centrally coordinated, have been successfully applied to real robots. The method of \cite{hoffmann2008decentralized} uses a centralized velocity obstacle (VO) occupancy map to perform collision avoidance for quadrotor helicopters. The work in \cite{schouwenaars2004decentralized} shows decentralized method for linear aircraft dynamics. A potential field implementation has been shown in simulations to work for non-holonomic agents, specifically fixed wing aircraft \cite{lalish2008decentralized}. Like potential field methods, di-polar navigation functions have been shown in simulations  to guarantee collision free paths for aircraft agents in the presence of centralized sensing uncertainty \cite{4738792}.

Important specifically to on-board, real-time sensing is the addition of noise to the velocity obstacle model. In \cite{snape2009independent}, the robots are enlarged by their one-sigma uncertainty ellipse to construct a conservative representation of the velocity obstacle. The work introduced in \cite{hennes2012multi} incorporates sensing uncertainty by increasing the size of an agent as a function of the noise on the position signal. This work has been extended in the CLAU  method which has been demonstrated on differential drive robots \cite{claes2012calu}. CLAU relies on a centralized model of the positions and velocities of agents using an uncertainty bounded by a particle filter.
 
On-board sensing limitations have been investigated in a variety of ways. Simulations of decentralized sensors with limited observation spaces have shown to not hinder collision avoidance given dynamic constraints \cite{dimarogonas2005decentralized}.  This topic is further explained for  holonomic vehicles with bounded velocity in \cite{loizu2004decentralized}.

In this paper we hypothesize that sensing uncertainty in reciprocal collision avoidance can lead to \emph{reciprocal dances}. Reciprocal dances have a limited theoretical backlog of work, but a few researchers have investigated it in both human crowd dynamics and agent simulation. First mentioned in 1971 \cite{goffman1971relations}, this phenomenon was shown in human experiments concerning movement through bottlenecks in \cite{hoogendoorn2005pedestrian}. A crowd collision model agrees with the hypothesis that miscommunication between agents leads to the non-optimal behavior \cite{feurtey2000simulating}. Viewing human miscommunication as a form of sensing uncertainty, we similarly hypothesize that reciprocal dances are a direct result of asymmetry in the sensed relative positions and velocities by pairs of agents, and provide a theoretical explanation for the phenomenon in the context of RCA in this paper.

\section{Reciprocal Collision Avoidance} \label{orca}
In this section we review the optimal reciprocal collision avoidance approach, and discuss qualitatively the impact the violation of basic assumptions has on the resulting motion.

\subsection{Optimal Reciprocal Collision Avoidance}
\begin{figure}
\centering
\includegraphics[width = \columnwidth]{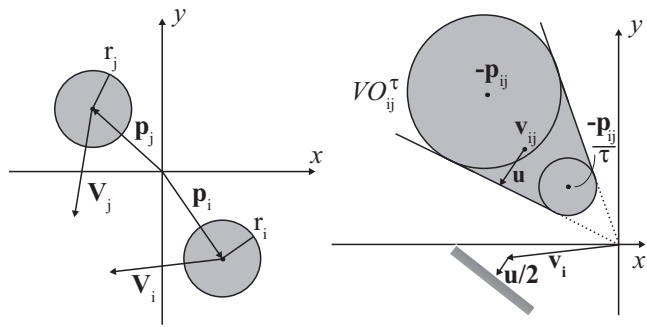}
\caption{Shown on the left is an example configuration for two robots, $ i$ and $j $, which will lead to the velocity obstacle shown on the right. The updated velocity for robot $ i $ is shown to be updated by one-half of $ \mathbf{u} $ as designed in ORCA.}
\label{Fig:nBody}
\vspace{-10pt}
\end{figure}

The ORCA algorithm is at its core a velocity obstacles method. Given two agents $i$ and $j$ sharing a workspace, the velocity obstacle $ \mathcal{VO}_{ij}^\tau $ of robot $i$ with respect to robot $j$ is a geometric representation of the set of \emph{relative} velocities that will result in a collision between robot i and j within $\tau $ time in the future. Each robot can be represented by a simple shape such as a disc of radius $ r $. For robot $ i $ to create a velocity obstacle with respect to robot $ j $ it must be able to determine \emph{relative} position $ \mathbf{p}_{ji} = \mathbf{p}_j - \mathbf{p}_i $ and the sum of the radii of the robots (see Fig.\ \ref{Fig:nBody}). Given a velocity obstacle, a collision is imminent before time $ \tau $ if
\begin{equation}
\mathbf{v}_{ij} \in \mathcal{VO}_{ij}^\tau,
\end{equation}
where ${v}_{ij}= v_i - v_j$, is the current relative velocity.

Referring to Fig. \ref{Fig:nBody}, ORCA determines the minimum change in relative velocity necessary to avoid collision represented by vector $ \mathbf{u} $. The reciprocal aspect of the algorithm assumes each robot will take equal responsibility to avoid collision, therefore, each robot changes their velocity by at least half of the required change thus changing the \emph{relative} velocity by at least the full correction. Robot $ i $ would then change its velocity by at least $ \frac{1}{2}\mathbf{u} $ such that $ \mathbf{v}_i $ is in the halfplane through $ \mathbf{v}_i + \frac{1}{2}\mathbf{u} $ perpendicular to $\mathbf{u}$. Symmetrically, robot $j$ changes its velocity by at least $ \mathbf{v}_j - \frac{1}{2}\mathbf{u} $

In all, a robot $i$ requires the relative position, the relative velocity, and its own absolute velocity to construct the halfplane of valid new velocities with respect to each robot $j$. It then selects a new velocity from the intersection of all halfplanes. The concept is naturally extended to 3-D.

\subsection{Non-Cooperative Agents and Exponential Convergence}
One key assumption for \emph{ORCA} to guarantee collision avoidance is that the agents sharing the workspace are cooperative and will perform the necessary reciprocal action to avoid collision. In practice, this assumption may be violated, for example, through loss of on-board tracking of another agent. The robustness of the algorithm allows for collision avoidance to still occur through exponential convergence to a collision free-path if only one robot is reacting. 

\begin{figure}
\centering
\includegraphics[width = \columnwidth]{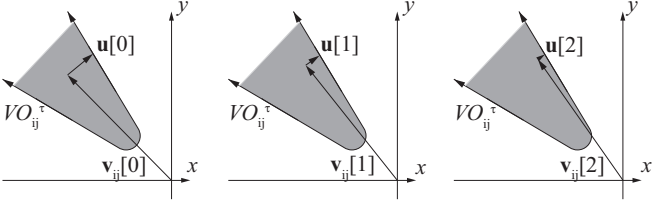}
\caption{Assuming an infinitesimally small time-step, the velocity obstacle does not change from each previous time-step to the next. Only the relative velocity changes towards a free velocity each time-step. The velocity obstacle in three consecutive time steps is shown where $ \mathbf{v}_{ij} $ converges exponentially towards a collision-free trajectory.}
\label{fig:exponential}
\vspace{-10pt}
\end{figure}

We show this using an extreme case in which there are two robots, $i$ and $j$, where robot $ i $ is using ORCA and robot $ j $ is a non-cooperative agent moving along a constant velocity $\mathbf{v}_j$ with no regard to robot $ i $. Robot $ i $ will create a velocity obstacle given a current configuration and, per \emph{ORCA}, update its velocity to avoid collision. However, robot $i$ is expecting the other agent to perform half the necessary action. Therefore, in the first adaptation, robot $ i $ is not controlled completely out of collision, but only makes half the necessary change. Assuming an infinitesimally small time step, in the next sensing-action cycle robot $i$ creates (almost) the same velocity obstacle, but the necessary change in relative velocity between the robots is half of what it was in the previous cycle, and again robot $i$ takes care of half of this necessary change. The velocity of $i$ converges to a collision-avoiding velocity according to the difference equation
\begin{align}
\mathbf{v}_i[t+\Delta t] & = \mathbf{v}_i[t] + \frac{1}{2}(\mathbf{v}^*_{ij} - \mathbf{v}_{ij}[t]) \nonumber \\
& = \mathbf{v}_i[t] + \frac{1}{2}(\mathbf{v}^*_{ij} - \mathbf{v}_{i}[t] + \mathbf{v}_j),
\end{align}
where $\mathbf{v}^*_{ij}$ is the relative velocity closest to the current relative velocity outside the velocity obstacle (see Fig. \ref{fig:exponential}). 

The difference equation solves to:
\begin{align}
\mathbf{v}_i[t] = (\mathbf{v}^*_{ij} + \mathbf{v}_j) - {2^{-t/\Delta t}}(\mathbf{v}^*_{ij} - \mathbf{v}_{i}[0] + \mathbf{v}_j). \label{eq:solutiondifferenceequation}
\end{align}
As $t \rightarrow \infty$, the velocity $\mathbf{v}_i$ of robot $i$ converges to  $(\mathbf{v}^*_{ij} + \mathbf{v}_j)$ and hence the relative velocity $\mathbf{v}_{ij}$ to $\mathbf{v}^*_{ij}$, which is a collision avoiding relative velocity. As can be seen from Eq.\  \eqref{eq:solutiondifferenceequation}, this convergence happens with an exponential rate. A slight overestimation of the robot's radii is therefore enough in most practical cases to ensure collision avoidance even if not all robots act as assumed.

The same mechanism is at play when the robots have dynamics constraints that prevent them from acting as assumed. For instance, if the robots are not able to change their velocity instantaneously, but can only control their acceleration, we may let the acceleration be controlled as:
\begin{align}
\mathbf{a}_i[t] = k_i \frac{1}{2}(\mathbf{v}^*_{ij} - \mathbf{v}_{ij}[t]),
\end{align}
for some control gain $k_i$. That is, the acceleration of robot $i$ is set proportional to the required change in velocity for robot $i$ as set by ORCA. This leads to the differential equation:
\begin{align}
\dot{\mathbf{v}}_i[t] = \mathbf{a}_i[t] = k_i \frac{1}{2}(\mathbf{v}^*_{ij} - \mathbf{v}_{i}[t] + \mathbf{v}_j),
\end{align}
for infinitesimally small time steps, which has as solution:
\begin{align}
\mathbf{v}_i[t] = (\mathbf{v}^*_{ij} + \mathbf{v}_j) - e^{-k_i t / 2}(\mathbf{v}^*_{ij} - \mathbf{v}_{i}[0] + \mathbf{v}_j).
\end{align}
Hence, in this case the relative velocity $\mathbf{v}_{ij}$ converges to the collision avoiding relative velocity $\mathbf{v}^*_{ij}$ at an exponential rate.

	\subsection{Sensing Uncertainty and Reciprocal Dances}
A second key assumption for ORCA to guarantee collision avoidance is that there is perfect symmetry between robot $i$ and $j$. More formally, the relative position $\mathbf{p}_{ji}$ and relative velocity $\mathbf{v}_{ij}$ as sensed by robot $i$ are the exact negative of the relative position $\mathbf{p}_{ij}$ and relative velocity $\mathbf{v}_{ji}$ as sensed by robot j. In a perfect world, this is the case by definition, and as a result we have that $\mathcal{VO}_{ij}^\tau = - \mathcal{VO}_{ji}^\tau$:
\begin{equation}
\mathbf{p}_{ji} = -\mathbf{p}_{ij},\; \mathbf{v}_{ij} = -\mathbf{v}_{ji} \; \rightarrow \; \mathcal{VO}_{ij}^\tau = - \mathcal{VO}_{ji}^\tau.
\end{equation}
As shown in Fig.\ \ref{Fig:nBody}, when the relative velocity $\mathbf{v}_{ij}$ lies to the left of the center line of the velocity obstacle $\mathcal{VO}_{ij}^\tau$ (as seen from the origin), then the ORCA halfplane is constructed such that robot $i$ will choose a velocity to pass robot $j$ on its left. By symmetry, the relative velocity $\mathbf{v}_{ji}$ as seen by robot $j$ lies to the left of the center line of $\mathcal{VO}_{ji}^\tau$ (again, seen from the origin), and robot $j$ will choose to pass robot $i$ on its left, leading to smooth collision avoiding motion by both robots. 

More formally:
Robot $ i $ selects a new velocity from its ORCA halfplane based on $ \mathcal{VO}_{ij}^\tau $ and robot $ j $ selects a new velocity from its ORCA halfplane based on $ \mathcal{VO}_{ji}^\tau $ such that:
\begin{align}
\mathbf{v}_i[t+\Delta t] &= \mathbf{v}_i[t] + \frac{1}{2}\mathbf{u}_i, \\
\mathbf{v}_j[t+\Delta t] &= \mathbf{v}_j[t] + \frac{1}{2}\mathbf{u}_j,
\end{align}
which when expressed as relative velocities can be seen as
\begin{align}
\mathbf{v}_{ij}[t+\Delta t] &= \mathbf{v}_{ij}[t] + \frac{1}{2}\left(\mathbf{u}_i - \mathbf{u}_j\right), \\
\mathbf{u}_i &= -\mathbf{u}_j, \;\; \text{given} \; \mathcal{VO}_{ij} = -\mathcal{VO}_{ji}, \\
\mathbf{v}_{ij}[t+\Delta t] &= \mathbf{v}_{ij}[t] + \mathbf{u}_i.
\end{align}
So, indeed, the new relative velocity $\mathbf{v}_{ij}[t+\Delta t]$ is a collision avoiding velocity.

However, in practice this symmetry is broken due to sensor noise on each of the robots. Let $\mathbf{v}_{ij}$ denote the true relative velocity of robot $i$ and $j$, then the sensed relative velocities $\tilde{\mathbf{v}}_{ij}$ by robot $i$ and $\tilde{\mathbf{v}}_{ji}$ by robot $j$ can be assumed to be:
\begin{align}
\tilde{\mathbf{v}}_{ij} & = \mathbf{v}_{ij} + \mathbf{m}_i, & \mathbf{m}_i & \sim \mathcal{N}(\mathbf{0}, M_i), \\
\tilde{\mathbf{v}}_{ji} & = -\mathbf{v}_{ij} + \mathbf{m}_j, & \mathbf{m}_j & \sim \mathcal{N}(\mathbf{0}, M_j),
\end{align}
where $\mathbf{m}_i$ and $\mathbf{m}_j$ are the sensing noise of robot $i$ and $j$ drawn from a Gaussian distribution with zero mean and variance $M_i$ and $M_j$, respectively. Similarly, the relative position may be sensed with noise, but we focus on the sensing uncertainty of the relative velocity here.

The result is that if the true relative velocity $\mathbf{v}_{ij}$ lies very close to the center line of $ \mathcal{VO}_{ij}^\tau $ (meaning that the robots are on a ``head-on'' collision-course), then the relative velocity $\tilde{\mathbf{v}}_{ij}$ sensed by robot $i$ may be to the right of the center line of $ \mathcal{VO}_{ij}^\tau $, while the relative velocity $\tilde{\mathbf{v}}_{ji}$ sensed by robot $j$ may be to the left of the center line of $ -\mathcal{VO}_{ij}^\tau $, or vice versa. This causes the robot $i$ to want to pass robot $j$ on its right, and robot $j$ to want to pass robot $i$ on its left, and the new velocities chosen by the robots are most likely anew on a collision course. This asymmetry may repeat a number of time-steps, giving rise to a \emph{reciprocal dance}.

\begin{figure}
\centering
\includegraphics[width = \columnwidth]{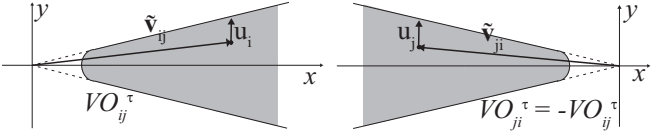}
\caption{Given two robots, $ i $ and $ j $, aligned along the global x-axis, a relative velocity obstacle can be created for each robot. The relative velocity obstacle is compared to the relative velocity to check for collision. Shown are two measured relative velocities which experience effects of uncertainty and are not symmetric. Given the asymmetry, if both relative velocities are on opposite sides of the velocity obstacle's centerline with respect to the origin, the robots will not avoid collision and undergo a reciprocal dance. Such an asymmetric measured relative velocity is shown.}
\label{Fig:reciprocalDance}
\vspace{-10pt}
\end{figure}

More formally, the lack of symmetry prevents the assumption that $ \mathbf{u}_i = -\mathbf{u}_j $. When $\tilde{\mathbf{v}}_{ij}$ and $\tilde{\mathbf{v}}_{ji}$ lie on the same side of the center lines of $ \mathcal{VO}_{ij}^\tau $ and $ -\mathcal{VO}_{ij}^\tau $, respectively (as seen from the respective agent), then $ \mathbf{u}_i \approx -\mathbf{u}_j $, and collision avoidance will smoothly occur given the discussion of the previous subsection. However, if $\tilde{\mathbf{v}}_{ij}$ and $\tilde{\mathbf{v}}_{ji}$ lie on opposite sides, then $ \mathbf{u}_i \not\approx -\mathbf{u}_j $, and a reciprocal dance will occur. More precisely, a reciprocal dance is expected to occur when
\begin{equation}
\mathbf{u}_i \cdot \mathbf{u}_j > 0,
\end{equation}
and the robots will choose a collision avoiding trajectory when 
\begin{equation}
\mathbf{u}_i \cdot \mathbf{u}_j < 0.
\end{equation}
For a deterministic system with perfect symmetry we have $\mathbf{u}_i = -\mathbf{u}_j$, so it is always the case that $\mathbf{u}_i \cdot \mathbf{u}_j < 0$. A set of relative velocity obstacles for robots $ i $ and $ j $ can be seen in Fig. \ref{Fig:reciprocalDance} where the sensing error in relative velocity will lead to a reciprocal dance.

It should be noted that the reciprocal dance is also a very human phenomenon.  When two people find themselves walking down a narrow corridor on a collision course, both people try to avoid collision by stepping to one side or the other. However, without communication there are times when both people move the same absolute direction, beginning a cyclic back and forth movement until eventually one person makes a move opposite of their opposer, allowing both to move on without collision. This same behavior was observed during our robotic experiments and can be explained by the presence of sensing noise in the context of the ORCA formalism.
We note that any approach introduced before to accommodate sensing uncertainty in reciprocal collision avoidance cannot prevent the occurrence of reciprocal dances without mutual coordination or communication between robots. 

Further analysis of this behavior as it relates to sensing uncertainty could lead to a better understanding of acceptable noise in sensing hardware. We would like to investigate this more extensively in future work. For this paper, however, the focus is on the functionality and viability of the algorithm.

\section{Implementation Details} \label{sensing}
We performed our experiments on Parrot AR.Drone 2.0 quadrotors both in an environment with motion capture equipment (to record ``ground truth'' data on the motion of the robots) and in environments without. These quadrotor helicopters were remote-controlled by human operators (i.e. the human operatators set their preferred velocity) and exhibit different dynamics than the idealized assumption that new velocities can be adopted instantaneously. Each quadrotor has an on-board forward facing camera to detect other quadrotors (each quadrotor carries a tag for identification) and a downward-facing camera to estimate its own absolute velocity.

\subsection{System Overview}\label{setup}

\begin{figure}
\centering
\includegraphics[width=\columnwidth]{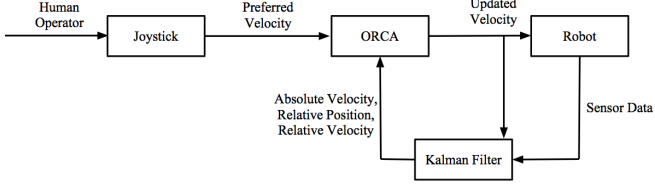}
\caption{Shown is a schematic overview of our experimental system. A human operator controls each robot through a joystick. That user-desired velocity is checked in ORCA for collision and if necessary, it is updated to be a collision-free velocity. ORCA receives the relative position, relative velocity, and individual absolute velocity information from the Kalman filter. The updated collision free velocity from ORCA is then sent to the robot's internal controller as well as to the Kalman filter. Sensor data from the actual robot is input to the Kalman filter.}
\label{Fig:blockDiagram}
\vspace{-10pt}
\end{figure}

Our system is set up as schematically shown in Fig.\ \ref{Fig:blockDiagram} for each quadrotor. Using a joystick, the human operator indicates the preferred velocity for the quadrotor it is operating. This preferred velocity is taken as input by ORCA, which, assuming knowledge of the relative positions and relative velocities with respect to other quadrotors, outputs a safe velocity to adopt. This safe velocity is sent to the on-board software of the quadrotor (we use the standard software on-board the AR Drone 2.0 to abstract our discussion away from issues of quadrotor control), which steers the motors on the quadrotor to adopt this velocity. Note that due to quadrotor dynamics, this velocity is not assumed instantaneously. This input velocity, together with sensing data of the other quadrotors on the imaging plane of the forward-facing camera is processed in a Kalman filter to estimate the relative positions and velocities of the other quadrotors. The absolute velocity of the quadrotor is estimated by the standard on-board software. This information is constantly fed to the ORCA module, which ``transforms'' the desired velocity of the operator to a velocity deemed safe by ORCA.

\subsection{ARDrone 2.0 - Parrot}
We use the Parrot AR.Drone 2.0, a generally available off-the-shelf quadrotor platform for our experiments. Operation is generally conducted with a software development kit (SDK) released from the manufacturer \cite{parrot_sdk}. Its sensors include:
				\begin{itemize}
					\item Nine degree-of-freedom IMU (composed of rate gyroscope, accelerometer, and magnetometer)
					\item Ultrasound and pressure sensors
					\item 30 Hz forward-facing camera
					\item 60 Hz downward-facing camera
				\end{itemize}
				
The AR.Drone stabilizes via an on-board PID controller \cite{bristeau2011navigation}, which acts on the orientation of the drone as a function of the onboard software's velocity estimate and the user input. Its onboard software also provides pixel coordinates and distance to a special ``tag'' marker attached to the robot as seen by the forward facing camera. The size of the tag is known by the internal SDK and can be used to estimate distance. We use this tag data in lieu of creating our own vision algorithm; this allows us to focus on the ORCA algorithm and its effectiveness.

\subsection{Kalman Filter for Relative Positioning}
As discussed above, for reciprocal collision avoidance, each quadrotor $i$ needs to know the relative position $\mathbf{p}_{ij}$ with respect to other quadrotors $j$, the relative velocity $\mathbf{v}_{ij}$ with respect to $j$, and its own absolute velocity $\mathbf{v}_i$. To keep track of these quantities, we implemented a Kalman filter for each quadrotor $i$ with a simplified dynamics model: 
	\begin{align}
	\dot{\mathbf{p}}_{ij} & = \mathbf{v}_i - \mathbf{v}_j, & & \forall(j \not= i) \label{eq:vel_p}\\
	\dot{\mathbf{v}}_i & = k(\mathbf{v}_i^* - \mathbf{v}_i), & & \\
	\dot{\mathbf{v}}_j & = \mathbf{m}_j, & & \forall (j \not=i),
	\end{align}
where $\mathbf{m}_j \sim \mathcal{N}(\mathbf{0},M)$.	Here, $\mathbf{v}_i^*$ is the new velocity for quadrotor $i$ as output by the reciprocal collision avoidance algorithm; since we do not know this quantity for any other quadrotor, we assume the evolution of the other quadrotor's velocities to resemble a random walk with variance $M$.

Through the on-board camera, each quadrotor $i$ measures the planar pixel coordinates $\tilde{\mathbf{b}}_j$ of other quadrotors $j$ in its imaging plane, as well as their distance $\tilde{d}_j$ (using the size of the tag on quadrotor $j$), and can deduce the other quadrotor's relative position using these polar coordinates. Also, a separate on-board, higher frequency Kalman filter keeps track of the full state of the quadrotor (velocity, orientation, angular velocity); this second Kalman filter provides ``measurements'' $\tilde{\mathbf{v}}_i$ of its velocity and $\tilde{R}_i$ of its orientation from world frame to camera frame, that we will use to define the measurement model of the relative positioning Kalman filter. The measurement model is of the general form $\mathbf{z} = h(\mathbf{x}) + \mathbf{n}$, with $\mathbf{n} \sim \mathcal{N}(\mathbf{0}, N)$, where $\mathbf{z}$ consists of $\tilde{\mathbf{v}}_i$ and $\tilde{\mathbf{b}}_j$ and $\tilde{d}_j$ for all $j \not= i$. The function $h$ is defined by:  
\begin{align}
\tilde{\mathbf{b}}_j & = \tilde{R}^T_i \mathbf{p}_{ij} f / (\tilde{R}^T_i \mathbf{p}_{ij})_z, & & \forall(j \not= i)\\
\tilde{d}_j & = \| \mathbf{p}_{ij} \|, & & \forall(j \not= i)\\
\tilde{\mathbf{v}}_i & = \mathbf{v}_i,
\end{align}
where $f$ is the focal length of on-board camera (assuming it is a pinhole camera), and $(\mathbf{p})_z$ denotes the $z$ component of vector $\mathbf{p}$. 

\subsection{ORCA}
For the reciprocal collision avoidance in our system, we use the standard implementation of 3-D ORCA as publicly available on http://gamma.cs.unc.edu/RVO2/. Each quadrotor helicopter is geometrically modeled in ORCA as an ellipsoid encompassing the quadrotor elongated along the z-axis to discourage pairs of quadrotors to fly in each other's downwash from the propellers. 
		
\section{Results} \label{results}
In this section, we report results of our experiments. Firstly, we quantify the accuracy of the estimation of relative position and relative velocity using the on-board sensors and the Kalman filter by comparison to ``ground truth'' data obtained from a motion capture system. Second, we report on over one hundred runs with a pair of quadrotors that were controlled by human operators to fly into one another, and qualitatively describe the observed motion. None of our more than one hundred runs resulted in a collision between the robots.
			\begin{figure}
			\centering
			\includegraphics[width=\columnwidth]{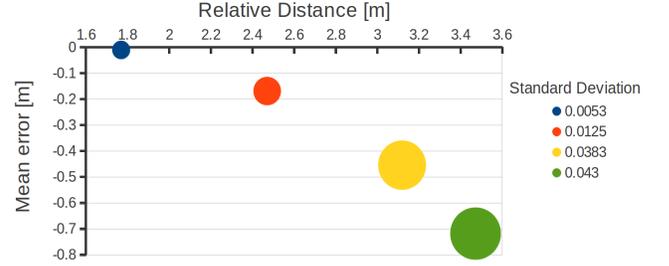}
			\caption{Quadrotors were flown in a straight-line trajectory by the user where ORCA was the sole method of avoiding collision. The graph plots error in the relative position measurement against the true relative position. As the robots come closer together, the sensing uncertainty exponentially converges to a small standard deviation with zero mean.}
			\label{fig:KfError}
			\vspace{-10pt}
			\end{figure}

			\begin{figure}[!h]
			\centering
			\includegraphics[width=\columnwidth]{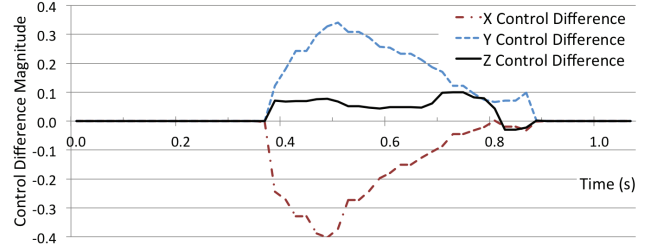}
			\caption{In this experiment, one quadrotor was flown on a colliding trajectory toward a stationary, non-cooperative quadrotor. The graph plots the difference between the original user input and the input given to the robot after adjustments by ORCA. Non-zero control difference means that a collision is detected and ORCA is changing the input.}
			\label{Fig:joyvscmd}
			\vspace{-20pt}
			\end{figure}

\subsection{Relative Position and Velocity Estimation}
Human operators flew two quadrotors repeatedly on colliding trajectories. The motion capture system was used to collect a true position and velocity of each robot during the testing. The estimated values of position and velocity were collected from each robot. To quantify the accuracy of the Kalman Filter, the true relative position and estimated relative position were compared, shown in Fig. \ref{fig:KfError}. The error associated with the estimated relative position is relatively large when an agent enters the viewing envelope at a far distance. As the agents approach each other, this error decreases in mean and standard deviation as shown. Given that the derivative of the relative position over time is used to estimate the relative velocity (see Eq. \ref{eq:vel_p}), the error in estimated relative velocity is directly coupled to the estimated relative velocity error. 
		
As can be seen from the graph, the standard-deviation of the error increases with the distance between the quadrotors. This is expected, as the quadrotors are estimating distance with respect to each other based on the size (number of pixels) of the tag of the quadrotor in the imaging plane. Assuming a constant standard-deviation in the error of the pixel count, far-away quadrotors with a low pixel-count will have a relatively large standard deviation in the distance estimate. The graph also shows an underestimate of the true distance between quadrotors, with a bias that grows with the distance. Although not intentional, the on-board sensing provides conservative estimates of the relative position for the purpose of collision avoidance. Given the structured relation of the bias as a function of distance, this bias can relatively easily be ``calibrated out'', but that was not deemed necessary for our purpose of collision avoidance.

	\subsection{Collision Avoidance Behavior}
	
To validate the collision avoidance behavior, two quadrotors were flown at each other along a global x-axis at a height of approximately two meters off the ground. The experiment consisted of two users simultaneously controlling their robot straight forward on a collision-course with the other robot. The user continued to control the robot forward and allowed ORCA to change the motion to a collision avoiding velocity. Over one hundred experiments of this type were completed. In these experiments, a reciprocal dance was observed in about $ 25\%$ of the trials, one non-cooperative agent was observed in $ 5\%$ of the trials, and the remainder were smooth collision avoiding trajectories. \emph{None} of the experiments conducted resulted in a collision. One can imagine a case in which neither agent is cooperative, and ORCA is therefore not implemented; this case would result in a collision. This case would be caused by hardware malfunctions in both agents. In all experimental trials, however, this case was never experienced.  A summary of these results can be seen in Table \ref{tab:Results}. Also, videos of some of the discussed experiments can be seen at \url{http://arl.cs.utah.edu/research/orca_quad/}.

\begin{table}
	\caption{Summary of Experimental Results}
	\label{tab:Results}
    \begin{tabular}{|c|c|c|c|}\hline
    Behavior & \# of Trials & \% of Trials & \# of Collisions \\ \hline    
    Reciprocal Dance & 28 & 25.9\% & 0 \\     \hline    
    One Non-cooperative & 6 & 5.5\% & 0 \\     \hline  
    Both Non-cooperative & 0 & 0.0\% & 0 \\     \hline  
    Collision Avoidance & 74 & 68.5\% & 0 \\     \hline \hline
    Total & 108 & 100 \% & 0 \\   \hline
    \end{tabular} 
    \vspace{-10pt}
\end{table}

Experiments were run in which the two quadrotors began within the sensing range of the forward-facing camera. Typically these trials resulted in an expected collision-avoiding motion where each robot moved symmetrically about the vector between them, i.e. both agents move to left of the center-line as seen from their perspective. As the two quadrotors approach each other ORCA begins to update the user-commanded velocity to avoid collision. ORCA avoids collision in this case by reducing the x-velocity slightly and inducing a y-velocity to move laterally. This action is shown in Fig. \ref{Fig:vel_error}. As the quadrotors exit the collision region, the output from ORCA converged to the user-desired velocity from the human operator and the quadrotors continue to fly forward. An example of the expected trajectory was shown in Fig. \ref{Fig:intro}.
	
\begin{figure}
	\centering
	\includegraphics[width = \columnwidth]{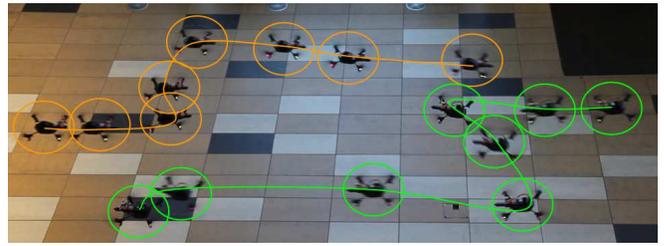}
	\caption{This experiment shows two quadrotors flown in a straight-line path by human operators. Due to sensing uncertainty, they undergo a reciprocal dance. Each robot initially flies in the same direction (up in the image) then, upon gaining a better relative velocity estimate, properly flies a non-colliding trajectory.}
	\label{fig:recipDance}
	\vspace{-10pt}
	\end{figure}

At times during these experiments, the error in the relative position measurement manifested in reciprocal dances as previously hypothesized. In these cases, both quadrotors moved in the same true direction leading to a colliding trajectory. Typically there was only one incorrect movement before both robots followed proper collision avoidance motions and moved symmetrically. These results were consistent with the hypothesis that sensing noise caused the reciprocal dances. As the robots move closer to each other, the sensing noise decreased and the robots were less likely to asymmetrically predict relative velocities. A sample reciprocal dance can be seen in Fig. \ref{fig:recipDance}.
	
In a small number of experiments, one of the quadrotors was unable to properly track the other quadrotor, leading to uncooperative behavior. Even in these experiments, no collisions were observed when a robot did not cooperate with the other. With no tracking information being received from the quadrotors on-board software, the Kalman filter could not provide a relative position estimate and, therefore, ORCA did not construct a valid velocity obstacle. As a result, the uncooperative robot remained on a straight-line trajectory as per the user's input. The second quadrotor, which did have tracking, properly avoided collision, further supporting our hypothesis that ORCA is robust to non-cooperative agents through exponential convergence.
	
\begin{figure}
	\centering
	\includegraphics[width = \columnwidth]{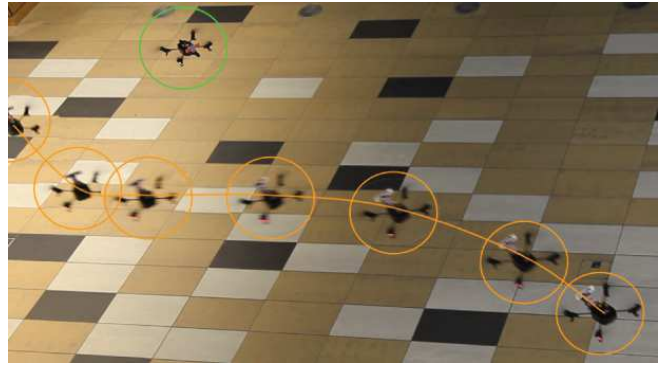}
	\caption{In this experiment, the top quadrotor, shown in orange, did not track the other quadrotor, shown in green.  As expected, the bottom quadrotor flies a non colliding velocity as a direct result of ORCA's ability to exponential converge to a collision-free trajectory given a non-cooperating, colliding, robot.}
	\label{fig:noncoop}
	\vspace{-10pt}
	\end{figure}

Experiments were also run to validate ORCA with noncooperative agents where a quadrotor without tracking was held stationary and a user-controlled quadrotor with tracking was flown directly at it along a straight-line trajectory. Such an experiment is shown in Fig. \ref{fig:noncoop}. Concurrent with other no-cooperative experiments, the robot who properly constructed the velocity obstacle took actions to avoid collision.
	
The dynamics of quadrotor helicopters were found to make ORCA less optimal, requiring it to rely on its ability to exponentially converge to collision free paths. A slightly larger bounding radius was the only requirement to allow for robust collision avoidance given exponential convergence to optimal collision avoidance. The fact that no collisions were observed in over one hundred experiments suggests that ORCA can provide reliable collision-avoidance even if the algorithm is not adapted specifically to account for the dynamics of the quadrotors or the sensing noise.
		
\section{Conclusions and Future Work}
	In this paper we have presented an analysis of Optimal Reciprocal Collision Avoidance on real robots using completely distributed acting and (on-board) sensing. With no centralized control or inter-robot communication, two quadrotors were able to successfully avoid collision in experimentation using built-in CMOS cameras and tracking algorithms. We analyzed the effects of violating key assumptions of ORCA and how they affect our system. We showed that sensing uncertainty can lead to reciprocal dances. These reciprocal dances were seen during several of the experiment trials. In addition, our work exhibits that non-cooperative agents can still avoid collision via exponential convergence. This was demonstrated at times when one of the two quadrotors did not properly construct the velocity obstacle, and was still able to avoid collision due to the other quadrotor's avoidance motion. 
	
	Our approach does have some drawbacks. The most notable is the use of the CMOS camera only allows tracking when the other robot is in a limited field of view. As the robots pass by each other and out of view, the Kalman filter uses the latest estimate and assumes this robot continues at that velocity. This assumption can obviously be violated due to the human control or external disturbances. Secondly, the estimated relative velocity from the Kalman filter is a derivative of the estimated relative position and as such can obtain very large values at times from noise in the position measurement. If this velocity estimates the robots flying quickly towards each other, it can cause the ORCA algorithm to behave as if the robots will collide imminently even when the robots are large distances apart. Often, this error is corrected in subsequent time steps, and the unnecessary motion will be small.
	
	For the future, we would like to further investigate the sensing uncertainty which leads to reciprocal dances. A better understanding of this dance could lead to a reciprocal collision avoidance algorithm which is more robust to sensing uncertainty. An algorithm which could tolerate large sensing uncertainty would allow for cheaper robotic sensors to be implemented, helping advance the field of diminishing cost in robotics. Another possible avenue of research is to investigate the improvement in collision avoidance capabilities from considering the dynamics of the system such as in \cite{Bareiss_2013}.

\bibliographystyle{ieeetran}
\bibliography{references}

\end{document}